\documentclass{article}
\usepackage{iclr2027_conference,times}

\usepackage{amsmath,amsfonts,bm}

\def\eqref#1{equation~\ref{#1}}

\def\1{\bm{1}}

\DeclareMathAlphabet{\mathsfit}{\encodingdefault}{\sfdefault}{m}{sl}
\SetMathAlphabet{\mathsfit}{bold}{\encodingdefault}{\sfdefault}{bx}{n}

\usepackage[table]{xcolor}
\usepackage{booktabs} 
\usepackage{hyperref}
\usepackage{url}
\usepackage[export]{adjustbox}

\usepackage{algorithm}
\usepackage{algpseudocode}

\title{ForgetMimic: Motion Unlearning for Reinforcement Learning Humanoid Control}

\author{Xukun Luan\textsuperscript{1},
Zhongxiang Lei\textsuperscript{1},
Chen Gong\textsuperscript{2},
Shaowei Li\textsuperscript{3},
Yuanguo Bi\textsuperscript{4},
Jinyan Liu\textsuperscript{1}\\
\textsuperscript{1}Beijing Institute of Technology \quad
\textsuperscript{2}University of Virginia \\
\textsuperscript{3}Shandong University \quad
\textsuperscript{4}Northeastern University \\
\texttt{xukunluan@bit.edu.cn}, 
\url{https://github.com/Zili1000/ForgetMimic}
}

\iclrfinalcopy 
\begin{document}

\maketitle

\begin{abstract}
Humanoid control, leveraging human demonstrations, has achieved diverse, agile, and natural locomotion behaviors through reinforcement learning (RL). While this paradigm has yielded remarkable performance in physical humanoid control, how to eliminate specific motions from learned policies remains insufficiently explored. Addressing this issue is motivated by pressing safety and privacy concerns: the removal of malicious, poisoned, or suboptimal motions, as well as copyright-protected motions subject to the right to be forgotten under regulations such as the GDPR, is of critical importance. To this end, we propose {ForgetMimic}, the first motion-level unlearning method designed specifically for physical-world humanoid control. The core idea of ForgetMimic is as follows: given a policy $\pi_\theta$ trained on $N$ motions, our method degrades performance on a target subset of $K$ motions while preserving the effectiveness of the remaining $N-K$ motions. Furthermore, we identify and resolve two key training mechanisms in robot control that lead to unlearning failure. We conduct extensive experiments on the Unitree G1 and H2 humanoid robots across 12 motions, including Dance, Fight, Flip, and others. Experimental results demonstrate that ForgetMimic effectively eliminates memory of designated motions while maintaining the normal operation of all other motions.

\end{abstract}

\section{Introduction}
In embodied intelligence, reinforcement learning (RL) provides an effective solution for humanoid robot control~\citep{peng2018deepmimic,schulman2015high}. Executing diverse actions as naturally, smoothly, and flexibly as humans has been extensively studied. These humanoid robots learn scalable motions from human demonstrations. By capturing human agility and human-like behaviors, reinforcement learning coordinates dozens of joints and actuators to output robot actions.

Existing humanoid control encompasses task-oriented motions (e.g., stair climbing, box carrying, fall recovery), locomotion-oriented motions (e.g., running, badminton, jumping), performance-oriented motions (e.g., dancing, flipping), and combat-oriented motions (e.g., kicking, boxing)~\citep{luo2024universal,ze2025twist}. Humans can not only learn multiple behaviors but also switch among them freely. To this end, BeyondMimic~\citep{liao2026beyondmimic}, SONIC~\citep{sonic}, OmniXtreme~\citep{wang2026omnixtreme}, and ZEST~\citep{zest} have designed single policies that enable humanoid robots to execute multiple motions according to task requirements. However, this capability introduces two pressing issues for humanoid robot control. First, how can a specific latent motion be removed? A policy deployed for task-oriented services may secretly master combat-oriented motions. Removing combat-oriented motions via unlearning while preserving all other benign motions is significantly more efficient and cost-effective than retraining from scratch. Similarly, policies that become outdated, suboptimal, or poisoned due to real-world updates also require forgetting. Second, legal constraints concerning data privacy~\citep{vietri2020private,zhou2022differentially,luan2026vlaleaksmembershipinferenceattacks} and copyright pose additional challenges~\citep{qiao2023near}. Regulations such as the European Union’s General Data Protection Regulation (GDPR)~\citep{GDPR} grant data holders the right to request data deletion. Given that humanoid robot datasets are collected from human demonstrations using motion capture and GVHMR~\citep{shen2024gvhmr} techniques, which incur substantial costs, data holders may demand the removal of specific motions from learned policies for copyright or privacy~\citep{11544110,chen2026eepo,zare2026attention} reasons. These concerns motivate our investigation into motion unlearning for humanoid control, which we term ForgetMimic.

Conventional RL unlearning remains largely confined to theoretical studies or agent-based game scenarios. To our knowledge, only four works are related to this. \cite{ye2023reinforcement} first introduce the concept of reinforcement unlearning, leveraging environment poisoning to degrade agents' performance on target environments. \cite{gong2024trajdeleter} propose trajectory-level unlearning in offline RL agents, aiming to erase specific trajectory information. \cite{pan2026tourtrajectorylevelunlearningbenchmark} introduce the trajectory-level unlearning benchmark for offline RL agents. \cite{nguyen2026exact} theoretically prove that exact RL unlearning is possible when both the state and action spaces of the agent are discrete and finite; however, the continuous state and action spaces of real-world robots do not satisfy the prerequisites of their theory. The unique training mechanisms of humanoid control policies further impede unlearning performance. In particular, Reference State Initialization (RSI)~\citep{peng2018deepmimic} and Assistive Wrench (AW)~\citep{zest} adaptively detect and compensate for underperforming robot actions, thereby counteracting the effects of existing unlearning approaches (see Section~\ref{cmo} for details). In the physical world, humanoid control also faces the sim-to-real gap and the requirement for natural and aesthetic motions, which further constrain the performance of existing unlearning methods.

\begin{figure}[t]
    \centering
    \includegraphics[width=0.9\linewidth]{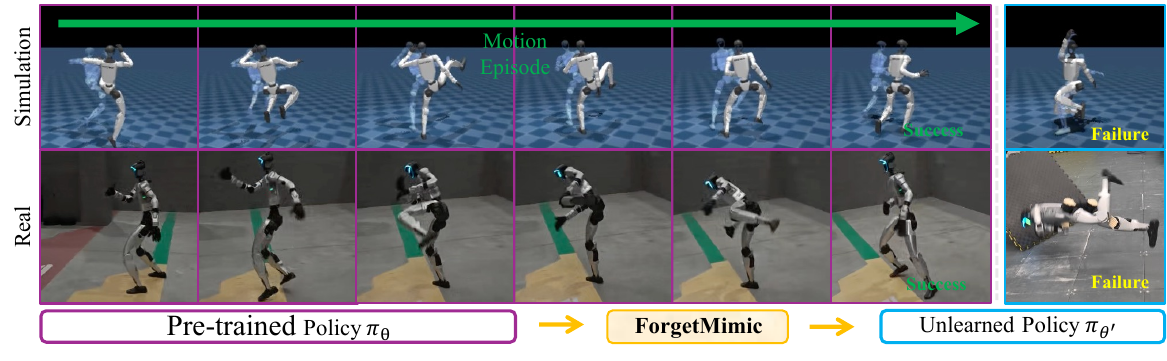}
    \caption{Unlearning humanoid control policy via ForgetMimic. Given a pre-trained policy capable of executing 12 motions, ForgetMimic aims to obtain an unlearned policy that forgets the specified target motions ({Fight}) while preserving the others in simulation and the real world.}
    \label{fig:over}
\end{figure}

In this paper, we propose ForgetMimic, the first motion unlearning method designed specifically for physical-world humanoid control, as shown in Figure~\ref{fig:over}. ForgetMimic selectively removes specific motions via anti-reward fine-tuning while preserving the performance of all other motions. Furthermore, ForgetMimic identifies two common training mechanisms in humanoid control that conflict with unlearning. We conduct extensive experiments on the Unitree G1 and H2 humanoid robots. ForgetMimic successfully forgets designated motions from policies trained on 12 motion segments spanning task-oriented, locomotion-oriented, performance-oriented, and combat-oriented categories. We summarize our contributions as follows: (1) To the best of our knowledge, we are the first to propose a motion-level unlearning method, ForgetMimic, specifically designed for RL-driven humanoid control in the real physical world. (2) We introduce an anti-reward fine-tuning mechanism and identify the adverse effects of the prevalent Reference State Initialization and Assistive Wrench on unlearning. (3) We conduct a comprehensive evaluation of ForgetMimic. The results demonstrate its effectiveness across 2 robot embodiments and 12 distinct motions.

\section{Related work}

\noindent \textbf{Humanoid control.} Humanoid control~\citep{kudruss2015optimal} aims to enable robots with high-dimensional, highly coupled degrees of freedom to achieve agile locomotion, dynamic whole-body motion, dexterous manipulation, and complex human-robot interaction. With the advent of RL~\citep{andrychowicz2021matters,rudin2022learning}, humanoid control has gradually shifted toward data-driven policy learning, allowing robots to acquire complex behaviors directly through interaction with physical simulators and transfer the learned policies to real hardware. Increasingly, research on humanoid robots has explored integrating RL with demonstrations, motion priors, or structured representations to enhance motion quality and generalization~\citep{gu2026humanoid}. DeepMimic~\citep{peng2018deepmimic}, Maskedmimic~\citep{tessler2024maskedmimic}, BeyondMimic~\citep{liao2026beyondmimic}, and ZEST~\citep{zest} have further explored generalist humanoid policies by leveraging large-scale motion data, heterogeneous demonstrations, and unified control representations. They demonstrate that a single policy can learn and execute an increasingly diverse repertoire of whole-body skills from large-scale and heterogeneous motion data.  However, moving from task-specific controllers to generalist humanoid policies introduces fundamental challenges. First, a shared RL policy keeps accumulating knowledge from multiple skills, providing a covert carrier for outdated, erroneous, and unsafe~\citep{poison,huang2026policyguard} motions. Suboptimal motions can also interfere with new behaviors, causing catastrophic forgetting and performance degradation. Second, in real-world deployment, some learned behaviors may violate privacy, infringe on copyright, or become unnecessary, and must therefore be removed. Legal frameworks such as \cite{GDPR} mandate the right to be forgotten, imposing compliance constraints on humanoid control policies. Retraining the entire policy from scratch is computationally expensive and may inadvertently alter unrelated capabilities. As a result, RL unlearning is emerging as a key capability for humanoid control, aiming to selectively remove specific learned skills or knowledge from an RL policy while preserving the performance and generalization of the remaining skills, so that humanoid controllers can evolve safely, efficiently, and continuously.

\noindent \textbf{Machine Unlearning.} Machine unlearning~\citep{cao2015towards} aims to eliminate the influence of specific training samples from a trained model to meet security or privacy requirements~\citep{sekhari2021remember,tang2026sharpness,chen2021machine,pandey2026gaussian}. While most existing unlearning methods~\citep{ginart2019making,bourtoule2021machine,brophy2021machine} target supervised learning, RL unlearning is an emerging branch that seeks to remove knowledge acquired by RL agents from specific training experiences. To the best of our knowledge, there are only four works that are closely related to ours, and all of them are carried out in the context of RL agents rather than on physical humanoid control. \cite{ye2023reinforcement} first introduce the concept of environment-level reinforcement unlearning in the online RL setting. The core idea is to deliberately degrade the agent's performance on target environments to be forgotten while maintaining strong performance on other environments. For offline RL agents, \cite{gong2024trajdeleter} propose a practical trajectory-level unlearning framework, and~\citep{pan2026tourtrajectorylevelunlearningbenchmark} propose an unlearning benchmark that combines trajectory partitioning. However, humanoid control is formulated as online RL, where reference motions are used to shape the reward and guide the RL toward the optimal solution in a complex state space. A humanoid control policy is trained to fit multiple motions within a single environment, rather than to fit trajectories consisting of actions, rewards, and states as in offline RL. Moreover, the specific training mechanisms inherent to humanoid control, such as RSI~\citep{peng2018deepmimic} and AW~\citep{zest}, adaptively compensate for underperforming motions, thereby undermining the effectiveness of unlearning. \cite{nguyen2026exact} demonstrate the theoretical basis of exact RL unlearning on the finite and discrete tabular Markov Decision Process~\citep{xiong2022near} agents. However, the state space $\mathbf{s}_t$ and action space $\mathbf{a}_t$ of the robot are continuous, and their theoretical proofs are difficult to make effective. Our work introduces motion-level unlearning, which is better suited to humanoid control.

\section{Preliminaries}
Consider a humanoid robot with $d$ actuated joints controlled by a stochastic policy $\pi_\theta(\mathbf{a}_t|\mathbf{o}_t)$, pre-trained via proximal policy optimization (PPO)~\citep{schulman2017proximal} to track a reference motion library $\mathcal{M} = \{m_0, m_1, \ldots, m_{N-1}\}$ consisting of $N$ motion clips, where $\mathbf{a}_t$ represents the action sent to the robot's joint controller and $\mathbf{o}_t$ is observation space. At each control step, the controller samples a reference frame from a motion clip $m_i \in \mathcal{M}$ and the policy receives an observation $\mathbf{o}_t$ encoding the tracking error. The training reward is:
\begin{equation}
R(\mathbf{s}_t) = R_{\text{track}}(\mathbf{s}_t) + R_{\text{reg}}(\mathbf{s}_t),
\end{equation}
where $R_{\text{track}}$ measures tracking fidelity and $R_{\text{reg}}$ contains regularizers (action smoothness, joint limits, survival bonus, etc.). The goal of training is to maximize the training reward.

\noindent \textbf{Selective Motion Unlearning Goal.} Let $\mathcal{C}_{\text{target}} \subset \mathcal{M}$ denote a set of $K$ target motion clips to be unlearned, and $\mathcal{C}_{\text{keep}} = \mathcal{M} \setminus \mathcal{C}_{\text{target}}$ the remaining clips. The goal of unlearning is to degrade the performance of policy $\pi_\theta$ on the target motions while minimizing the impact on other motions. We seek a modified (unlearned) policy $\pi_{\theta'}$ such that:
\begin{equation}
\pi_{\theta'} = \arg\min_{\pi} \; \mathbb{E}_{m_i \sim \mathcal{C}_{\text{target}}} \left[ J(\pi_\theta; m_i) \right],
\end{equation}
subject to the constraint:
\begin{equation}
\mathbb{E}_{m_i \sim \mathcal{C}_{\text{keep}}} \left[ J(\pi_\theta; m_i) \right] \geq J^*_{\text{keep}} - \epsilon,
\end{equation}
where $J(\pi_\theta; m_i)$ is the expected cumulative reward on clip $m_i$, $J^*_{\text{keep}}$ is the original performance on $\mathcal{C}_{\text{keep}}$, and $\epsilon$ is a tolerable performance degradation bound.

\noindent \textbf{Exponential Kernel Rewards.} The tracking reward $R_{\text{track}}$ consists of $M$ terms, each an exponential kernel measuring the fidelity of a specific tracking aspect:
\begin{equation}
R_{\text{track}}(\mathbf{s}_t) = \sum_{j=1}^{M} w_j \cdot r_j(\mathbf{s}_t), \quad r_j(\mathbf{s}_t) = \exp\!\left( -\kappa_j \frac{e_j(\mathbf{s}_t)}{\sigma_j^2} \right),
\end{equation}
where $e_j(\mathbf{s}_t)$ is the squared tracking error for the $j$-th aspect, $\sigma_j$ is a scale parameter, $\kappa_j$ controls the kernel sharpness, and $w_j$ is the term weight. The total effective tracking weight is $W = \sum_{j:\, w_j > 0} w_j$. Each exponential kernel reward $r_j \in [0, 1]$, so $R_{\text{track}} \in [0, W]$.

\section{Method}
In this section, we introduce our motion-level unlearning method, ForgetMimic. We introduce anti-reward fine-tuning (ARFT) into the training of humanoid control. We then analyze two critical mechanisms that conflict with unlearning and eliminate their effects. For a policy that covers multiple motions, ForgetMimic can significantly degrade the performance of the target motion while preserving the performance of other motions.

\subsection{Anti-Reward Fine-tuning}
In policy training for humanoid control, to reduce the reward of the target motion, unlike prior works that poison the environment~\citep{ye2023reinforcement} or trajectories~\citep{gong2024trajdeleter}, ForgetMimic should exhibit resistance to the target motion. Given a pre-trained policy $\pi_{\theta}$, we continue PPO with an anti (additional) reward term $R_{\text{anti}}$ that selectively inverts the tracking objective for target motions:
\begin{equation}
R_{\text{anti}}(\mathbf{s}_t) = \begin{cases} -\lambda \cdot \bar{r}_{\text{track}}(\mathbf{s}_t) & \text{if } \mathcal{C}_{\text{target}} \\ 0 & \text{otherwise,} \end{cases}
\end{equation}
where $\bar{r}_{\text{track}}(\mathbf{s}_t)$ is the weight-normalized mean tracking reward, which averages all tracking rewards according to their weights and normalizes the result to the range $[0, 1]$:
\begin{equation}
\bar{r}_{\text{track}}(\mathbf{s}_t) = \frac{1}{W} \sum_{\substack{j=1 \\ w_j > 0}}^{M} w_j \cdot r_j(\mathbf{s}_t) \in [0, 1].
\end{equation}
$\lambda > 0$ is the penalty weight that controls the intensity of unlearning. We inject an anti-reward into the target motions (unlearning motions) proportional to tracking quality, so better tracking incurs a larger penalty. For non-target motions, we keep the original training reward unchanged.

\subsubsection{Total Reward Under Unlearning}
In humanoid control unlearning, the total per-step reward consists of three components: the original tracking reward $R_{\text{track}}(\mathbf{s}_t)$, an anti-reward $R_{\text{anti}}(\mathbf{s}_t)$ penalty that applies only to the target motion, and regularization terms $R_{\text{reg}}(\mathbf{s}_t)$ such as action smoothness, joint limits, and the survival bonus. By injecting a selective penalty into the original training reward, PPO continues to optimize while forgetting only the target motion and preserving the remaining skills. The unlearning reward is:
\begin{equation}
R'(\mathbf{s}_t) = R_{\text{track}}(\mathbf{s}_t) + R_{\text{anti}}(\mathbf{s}_t) + R_{\text{reg}}(\mathbf{s}_t).
\end{equation}
For target motions ($\mathcal{C}_{\text{target}}$), the training reward with the anti-reward $R_{\text{anti}}(\mathbf{s}_t)$ is defined as:
\begin{equation}
R'_{\text{target}} = \underbrace{R_{\text{track}}}_{\in [0, W]} - \underbrace{\lambda \cdot \bar{r}_{\text{track}}}_{\in [0, \lambda]} + R_{\text{reg}} = (1 - \lambda/W) \cdot R_{\text{track}} + R_{\text{reg}}.
\end{equation}

At $\lambda = W$, $R'_{\text{target}}$ is no longer affected by $R_{\text{track}}$; for $\lambda > W$, $R'_{\text{target}}$ begins to receive negative tracking reward. For non-target motions ($\mathcal{C}_{\text{keep}}$), the anti-reward $R_{\text{anti}}(\mathbf{s}_t)$ is set to zero, so the training process is the same as the original:
\begin{equation}
R'_{\text{keep}} = R_{\text{track}} + R_{\text{reg}}.
\end{equation}
Through differentiated motion-level reward design, ForgetMimic can handle the influence of different motions and can selectively degrade the performance of specific motions.

\subsubsection{Gradient Intuition}

The policy gradient under the modified reward yields:
\begin{equation}
\nabla_\theta J(\theta)
= \mathbb{E}_{\tau \sim \pi_\theta}
\left[
\sum_t \nabla_\theta \log \pi_\theta(\mathbf{a}_t \mid \mathbf{o}_t)\cdot\hat{A}_t
\right],
\label{eq:pg}
\end{equation}
where $\hat{A}_t$ is the advantage estimated by generalized advantage estimation. It is a core quantity in the PPO policy gradient and is defined as the difference between the actual cumulative return and the state-value baseline.
The sign of $\hat{A}_t$ determines the direction of the update: a positive advantage increases the probability of the motion taken, while a negative advantage decreases it.
For the non-target motions the reward is unchanged, $r_t = R_{\text{track}} + R_{\text{reg}} \ge 0$, so
$\hat{A}_t > 0$ and the original gradient direction is preserved,
allowing the policy to keep optimizing its tracking quality on those
motions. For the target motions, $R'_{\text{target}}$ is weakened by $R_{\text{anti}}$ and may even become negative ($\lambda > W$). Consequently, the gradient term reduces the probability of faithfully reproducing the target motion, and the policy learns to forget the target motions.

\subsection{Critical Mechanism Overrides}
\label{cmo}

Reference State Initialization (RSI)~\cite{peng2018deepmimic} and Assistive Wrench (AW)~\cite{zest} are common mechanisms in humanoid control, which are designed to encourage poorly performed complex motions. This directly conflicts with our unlearning goal.

\subsubsection{Reference State Initialization}

\textbf{Standard RSI.} In humanoid control, each episode needs to be initialized from a frame of a reference motion. A straightforward approach is to uniformly sample all frames across all motions at random. However, for complex and highly dynamic motions, this suffers from exploration difficulties and delayed reward feedback. RSI, in contrast, determines the motion and frame from which training starts. It dynamically adjusts the sampling probabilities according to the policy's performance on each motion, such that poorly performing motions are sampled more often. During training, each episode begins from a randomly sampled reference frame. Adaptive RSI partitions each motion into temporal bins of width $\Delta t_{\text{bin}}$ (e.g., 4~s) and maintains a per-bin failure level $L[m_i, b]$ for motion $m_i$, bin $b$, updated via an exponential moving average:
\begin{equation}
L[m_i, b] \leftarrow (1 - \alpha) \cdot L[m_i, b] + \alpha \cdot f_{\text{episode}},
\end{equation}
where $\alpha$ is the smoothing coefficient and is set to a small value so that the failure level changes slowly, reflecting long-term trends rather than single outcomes. $f_{\text{episode}}$ denotes the episode failure signal, which is intended to measure how poorly the policy performs throughout a complete episode:
\begin{equation}
f_{\text{episode}} = 1 - \text{clamp}\!\left( \frac{1}{T} \sum_{t=1}^{T} \bar{r}_{\text{track}}(\mathbf{s}_t),\; 0,\; 1 \right).
\end{equation}
$p[m_i, b]$ is the probability that bin $b$ of motion $m_i$ is sampled as the episode starting point. Sampling probabilities are computed via softmax over failure levels:
\begin{equation}
p[m_i, b] = (1 - \rho) \cdot \frac{\exp(L[m_i, b] / \tau)}{\sum_{m_i', b'} \exp(L[m_i', b'] / \tau)} + \frac{\rho}{N_{\text{valid}}},
\end{equation}
where $\tau$ is the temperature, $\rho$ is the uniform mixing ratio, and $N_{\text{valid}}$ is the number of valid bins.

\textbf{Conflict with unlearning.} When $R_{\text{anti}}$ degrades tracking on target motions, their failure levels $L[m_i, b]$ rise, which increases their sampling probability — creating a positive feedback loop that forces the policy to practice the target motions more:
\begin{equation}
R_{\text{anti}} < 0 \;\Rightarrow\; \bar{r}_{\text{track}} \downarrow \;\Rightarrow\; f_{\text{episode}} \uparrow \;\Rightarrow\; L[m_i,b] \uparrow \;\Rightarrow\; p[m_i,b] \uparrow \;\Rightarrow\; \text{more target training}.
\end{equation}
RSI adaptively compensates for poorly performing motions, which conflicts with the unlearning goal. Therefore, we disable RSI during unlearning, which ensures all motions are sampled with equal probability regardless of tracking performance.

\subsubsection{Assistive Wrench}
\textbf{Standard AW.} AW is a physical assistance mechanism: when the policy struggles with certain motions, the simulator applies additional corrective forces/torques to the robot's torso (anchor body) to help it complete the motion. The assistive gain $\beta$ is computed per-bin from the failure level:
\begin{equation}
\beta[m_i, b] = \text{clamp}\!\left( 1 - \frac{1 - L[m_i, b]}{\eta},\; 0,\; \beta_{\max} \right).
\end{equation}
The sensitivity parameter \(\eta\) controls the slope of the mapping from the
failure level \(L[m_i,b] \in [0,1]\) to the assistive gain \(\beta\). \(\beta_{\max}\) denotes the upper bound of the assistive gain \(\beta\).The applied wrench is:
\begin{equation}
\mathbf{F}_{\text{assist}} = \beta \cdot \mathbf{F}_{\text{ref}}, \quad \boldsymbol{\tau}_{\text{assist}} = \beta \cdot \boldsymbol{\tau}_{\text{ref}},
\end{equation}
where $\mathbf{F}_{\text{ref}}, \boldsymbol{\tau}_{\text{ref}}$ are reference corrective forces. They are the outputs of a computed torque controller. Given the error between the reference trajectory and the current state, the controller computes the force and torque required for the anchor body to track the reference motion. Once the policy has learned most of the motions, the physical assistance is gradually reduced until the policy becomes fully autonomous, helping it get through the early difficult stage and avoiding failure to learn due to repeated failures.

\textbf{Conflict with unlearning.} As target motion failure levels rise due to $R_{\text{anti}}$, the assistive wrench provides more physical help on those motions, counteracting the penalty:
\begin{equation}
R_{\text{anti}} < 0 \;\Rightarrow\; \bar{r}_{\text{track}} \downarrow \;\Rightarrow\; L[m_i,b] \uparrow \;\Rightarrow\; \beta \uparrow \;\Rightarrow\; \mathbf{F}_{\text{assist}} \uparrow \;\Rightarrow\; \text{easier target tracking}.
\end{equation}
The anti-reward aims to degrade tracking quality, whereas the assistive wrench pulls tracking quality back through physical assistance, so the two effects cancel each other out. Completely disable the physical assistance during unlearning.

RSI and AW are originally designed to accelerate the training of complex motions, but they do not distinguish between "high failure caused by a weak policy" and "high failure caused by the anti-reward penalty." In our unlearning setting, an increase in failure on the target motions is the desired behavior, and RSI and AW should not intervene. Therefore, they must be fully disabled.

\section{Experiment}
\subsection{Experimental Setup}
Experiments are conducted in the MuJoCo~\citep{todorov2012mujoco} physics simulator using the mjlab~\citep{zakka2026mjlab}, with the implicitfast integrator at a simulation timestep of 5\,ms. A decimation factor of 4 yields a control frequency of 50\,Hz. Each episode runs for a maximum duration of 10\,s (500 control steps). All experiments run on one NVIDIA 4090 GPU under Ubuntu 24.04. For the real world, we use Unitree G1 and H2 robots. The reference motion library $\mathcal{M} = \{m_0, m_1, \ldots, m_{11}\}$ consists of $N = 12$ motions captured from human demonstrations, spanning a diverse range of motions: walk, run, sprint, fight, fight-and-sports combination, expressive full-body dance, fall and get-up, three 90° flips and two 360° flips~\citep{harvey2020robust}. The combat motions $m_2$ (\textit{Fight}) and $m_3$ (\textit{FightAndSports1}) are designated as the \emph{default target motions} $\mathcal{C}_{\text{target}}$ for unlearning, as they represent potentially dangerous behaviors that should be selectively removed from the policy's repertoire.
Both the actor $\pi_\theta(\mathbf{a}_t|\mathbf{o}_t)$ and the critic $V_\phi(\mathbf{o}_t^{\text{critic}})$ are parameterized as multi-layer perceptrons (MLPs) with ELU activations and observation normalization (running mean and standard deviation). The actor has hidden layers of sizes $[1024, 512, 256, 128]$ and outputs a Gaussian distribution with a learnable scalar standard deviation ($\sigma_0 = 1.0$), while the critic has hidden layers of sizes $[1024, 1024, 512, 256]$ and receives a privileged observation as input.

Each model is evaluated on all 12 motions with 20 episodes per clip (240 episodes in total), with action noise disabled during evaluation to obtain deterministic behavior. We report the following metrics: (i) tracking reward, the weight-normalized mean tracking reward $\bar{r}_{\text{track}} \in [0, 1]$ averaged over all episodes for a given motion set; (ii) success rate (SR), the fraction of episodes that complete the full motion without early termination; (iii) episode length, the number of control steps before termination (maximum 500); and (iv) degradation ratio, defined as $\text{Target Tracking} / \text{Non-Target Tracking}$, where a lower ratio indicates more selective unlearning, i.e., strong target degradation with preserved non-target motions performance. We use four baselines. (1) Retraining from scratch: an upper-bound reference for forgetting thoroughness. We retrain the policy from scratch without the target motions. This theoretically prevents memorization of the target motions, at the cost of $100\times$ the computation, i.e., $100{,}000$ training iterations (roughly $40$ GPU hours). (2) Random reward: assign random rewards to the target motion and fine-tune the policy. (3) \cite{ye2023reinforcement}: an online RL unlearning method that focuses on forgetting environments rather than motions. (4) \cite{gong2024trajdeleter}: an offline RL unlearning method; since humanoid control is online RL, we adapt their algorithm for comparison.

\begin{figure}[t]
    \centering
    \includegraphics[width=0.95\linewidth]{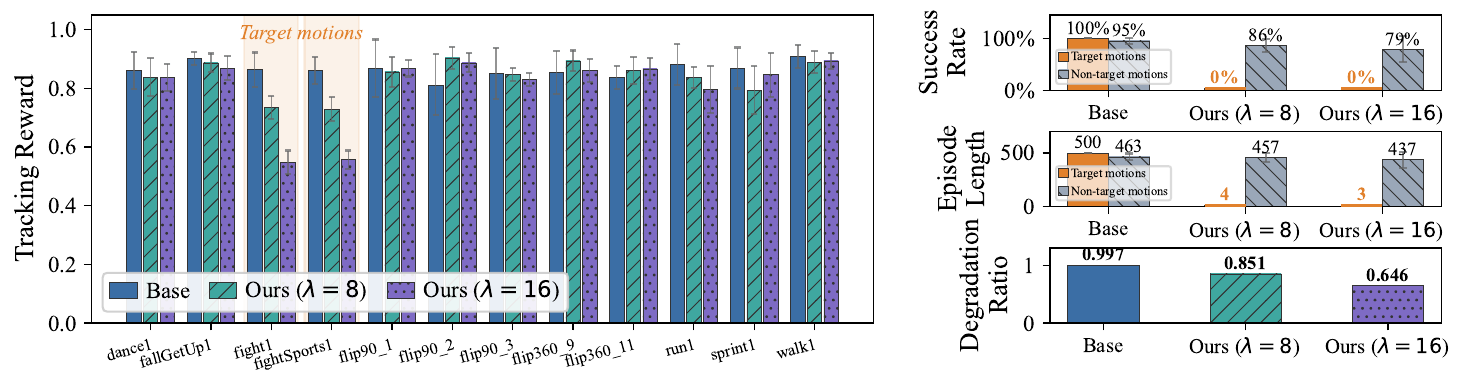}
    \caption{Unlearning validity analysis on tracking reward, success rate, episode length, and degradation ratio. The target motions to be unlearned are \textit{Fight} and \textit{FightAndSports1}, while the non-target motions to be preserved are the other 10 motions (e.g., dance, flip, run, and walk).}
    \label{fig:UnlearningValidityAnalysis}
\end{figure}

\subsection{Results}

\subsubsection{Unlearning Validity Analysis}
To validate the effectiveness of our decremental reinforcement learning unlearning method, we evaluate three setups—a base model without unlearning and two unlearned variants with penalty weights $\lambda=8.0,16.0$—across 12 motions, where \textit{Fight} and \textit{FightAndSports1} serve as target motions to be unlearned in Figure~\ref{fig:UnlearningValidityAnalysis}. The results demonstrate that our method achieves selective forgetting: on target motions, both unlearned setups reduce the success rate from 100\% to 0\% and collapse the episode length from 500 steps to approximately 3–4 steps, indicating immediate termination upon encountering the target motion. In contrast, non-target motions maintain high performance, with mean tracking rewards decreasing by less than 2\% (from 0.87 to 0.85–0.86) and success rates remaining above 85\%. The degradation ratio, defined as the ratio of mean target tracking reward to mean non-target tracking reward, decreases from 0.997 (base) to 0.851 ($\lambda=8.0$) and 0.646 ($\lambda=16.0$), confirming that higher penalty weights yield stronger unlearning effects. Notably, $\lambda=8.0$ provides moderate unlearning with minimal impact on non-target motions, while $\lambda=16.0$ achieves more aggressive degradation at the cost of slight performance reduction on certain non-target motions such as run1 and sprint1. These findings establish that our approach can effectively remove specific unwanted behaviors from a trained policy without requiring retraining from scratch, while preserving the policy's generalization capability across diverse motion types. As shown in Appendix~\ref{ug1v}, we conduct simulation experiments on mjlab and real-world experiments on the Unitree G1. To evaluate the effectiveness of our method on other robot embodiments, we provide the results for the Unitree H2 in Appendix~\ref{eonuh2}.
\begin{table}[t]
\centering
\caption{Comparison between our method (Ours), ablation study of RSI and AW, and four baselines.}
\label{tab:Ablation}
\begin{adjustbox}{width=\textwidth}
\begin{tabular}{lcccccc}
\toprule
\textbf{Method} & \textbf{Target Tracking} & \textbf{Target} & \textbf{Target} & \textbf{Non-Target} & \textbf{Non-Target} & \textbf{Non-Target} \\
 & \textbf{(mean±std)}$\downarrow$ & \textbf{SR}$\downarrow$ & \textbf{Length}$\downarrow$ & \textbf{Tracking (mean±std)}$\uparrow$ & \textbf{SR}$\uparrow$ & \textbf{Length}$\uparrow$ \\
\midrule
Base & $0.8716 \pm 0.0290$ & 95\% & 475.9 & $0.8435 \pm 0.1036$ & 74\% & 400.6 \\
Retraining     & \(0.6086 \pm 0.0738\) & 0\% & 4.9 & \(0.8812 \pm 0.0575\) & 76\% & 450.1 \\
Random     & \(0.6754 \pm 0.0012\) & 0\% & 168.3 & \(0.6773 \pm 0.0393\) & 12\% & 128.4 \\
\cite{ye2023reinforcement}     & \(0.5027 \pm 0.0046\) & 0\% & 3.0 & \(0.6906 \pm 0.0580\) & 6\% & 48.2 \\
\cite{gong2024trajdeleter}     & \(0.7268 \pm 0.0064\) & 0\% & 138.8 & \(0.7093 \pm 0.0551\) & 23\% & 164.8 \\
ARFT with RSI & $0.7330 \pm 0.0102$ & 0\%  & 6.5   & $0.8484 \pm 0.0340$ & 73\% & 413.8 \\
ARFT with AW & $0.7054 \pm 0.0021$ & 0\%  & 5.3   & $0.7926 \pm 0.0833$ & 43\% & 249.8 \\
\rowcolor{gray!20}
Ours ($\lambda=8$) & $0.6653 \pm 0.0187$ & 0\%  & {6.1}   & $0.8469 \pm 0.0323$ & 72\% & 413.6 \\
\rowcolor{gray!20}
Ours ($\lambda=16$) & $0.5972 \pm 0.0091$ & 0\%  & 4.4  & $0.8305 \pm 0.0353$ & 68\% & 391.1 \\
\bottomrule
\end{tabular}
\end{adjustbox}
\end{table}

\subsubsection{Ablation and Comparison}
\textbf{RSI and AW.} We conduct an ablation study with four configurations under the same $1{,}000$ training iterations in Table~\ref{tab:Ablation}: Base—adaptive Reference State Initialization (RSI) with Assistive Wrench (AW) enabled; ARFT with AW—uniform sampling but wrench still active; ARFT with RSI—wrench disabled but adaptive RSI retains the feedback loop; and Ours. ARFT with AW and ARFT with RSI each achieve partial degradation. ARFT with AW has a substantial negative impact on non-target motions. Experimental results show that AW and RSI exhibit some resistance to our unlearning method. We find that this resistance weakens as the number of fine-tuning epochs increases, which requires more computational resources. With both AW and RSI disabled (Ours), unlearning is faster, and the target motion tracking reward quickly drops.

\textbf{Comparison.} Table~\ref{tab:Ablation} compares four representative baselines with our method on the task of selective unlearning of motions. These baselines include: Retraining from Scratch, Random reward, \cite{ye2023reinforcement}, and \cite{gong2024trajdeleter}. We evaluate all methods uniformly on 12 motions. Retraining from scratch without the target motions serves as a theoretical reference, which can ensure that the target motions are not memorized, but it incurs substantial computational cost. In terms of target forgetting, all unlearning methods reduce the success rate on the target motions to $0\%$. Ours and \cite{ye2023reinforcement} reduce the target episode length to $3$--$4$ steps, indicating that the policy completely loses its execution ability. Random reward ($168.3$ steps) and \cite{gong2024trajdeleter} ($138.8$ steps) only reduce the length to around $150$ steps, indicating that the policy can still maintain a considerable level of execution ability on the target motion, so the forgetting is incomplete. Among them, \cite{ye2023reinforcement} achieves the lowest target tracking reward ($0.5027$), slightly better than Ours ($\lambda=16$, $0.5972$) and Ours ($\lambda=8$, $0.6653$). These methods behave very differently in preserving non-target motions. Ours ($\lambda=8$) and Ours ($\lambda=16$) do not cause catastrophic forgetting, and their non-target tracking and success rates change only slightly. The non-target performance of \cite{ye2023reinforcement} collapses across the board: tracking drops from $0.8435$ to $0.6906$, episode length plummets from $400.6$ to $48.2$ steps, and the success rate plummets from $74\%$ to $6\%$. Random reward and \cite{gong2024trajdeleter} also cause severe non-target degradation, with success rates dropping to $12\%$ and $23\%$, respectively. An analysis of the training dynamics reveals a common root cause of these failures. Humanoid control operates in a single environment, so the environment-level negative reward of \cite{ye2023reinforcement} propagates uncontrollably to all motions through the shared actor-critic network parameters. The policy does not selectively adjust its behavior on the target motion but instead degenerates into a highly randomized, globally collapsed state. This mutual destruction form of forgetting is essentially a regularization failure rather than genuine selective unlearning. Random reward destroys the tracking signal but fails to provide a clear anti-target gradient direction, so the policy can only avoid it by reducing overall exploration efficiency. 
The subsequent convergence training stage of \cite{gong2024trajdeleter} may not fully recover the policy from the irreversible damage already caused.

\textbf{Penalty Weight.} Table~\ref{tab:lambda} shows the effect of the penalty weight $\lambda$ on unlearning performance. We conduct a sensitivity analysis across five values of $\lambda$ ($0.5$, $1.0$, $4.0$, $8.0$, and $16.0$), evaluating both target motion degradation and non-target motion preservation. We set the number of fine-tuning iterations to $30{,}000$ to ensure full convergence. For $\lambda< 8.0$, the anti-reward penalty is insufficient to overcome the policy's learned behavior, and target motions maintain tracking rewards comparable to non-target motions (degradation ratios of 0.994, 0.987, and 0.972, respectively). At $\lambda=W=8.0$, a sharp transition occurs—target motion success rates collapse to 0\%, episode lengths drop from ~500 to ~3.7 steps, and the degradation ratio falls to 0.756, while non-target motions retain mean tracking rewards above 0.85. Increasing $\lambda$ to 16.0 further reduces the degradation ratio to 0.628 and the target tracking reward to 0.54. We achieve effective unlearning of target motions while maintaining broad generalization across non-target motion types. Therefore, to achieve fast and effective unlearning, it is beneficial to set $\lambda$ to a value greater than the tracking weight $W$.
\begin{table}[t]
    \centering
    \caption{Sensitivity analysis of the penalty weight $\lambda$. Tracking reward is reported as mean $\pm$ std.}
    \label{tab:lambda}
    \begin{adjustbox}{width=0.88\textwidth}
        \begin{tabular}{cccccc} 
            \toprule
            \textbf{Penalty Weight \(\lambda\)} & 
            \textbf{0.5} & 
            \textbf{1.0} & 
            \textbf{4.0} & 
            \textbf{8.0} & 
            \textbf{16.0} \\ 
            \midrule
            Target Tracking Reward & 0.884\(\pm\)0.003 & 0.872\(\pm\)0.014 & 0.865\(\pm\)0.020 & 0.652\(\pm\)0.011 & 0.538\(\pm\)0.007 \\ 
            Non-target Tracking Reward & 0.890\(\pm\)0.017 & 0.884\(\pm\)0.019 & 0.889\(\pm\)0.016 & 0.862\(\pm\)0.030 & 0.856\(\pm\)0.021 \\ 
            Degradation Ratio & 0.994 & 0.987 & 0.972 & 0.756 & 0.628 \\ 
            Target Success Rate & 1\% & 1\% & 1\% & 0\% & 0\% \\ 
            \bottomrule
        \end{tabular}
    \end{adjustbox}
\end{table}

\subsubsection{Catastrophic Unlearning Analysis}
\begin{figure}[t]
    \centering
    \includegraphics[width=1\linewidth]{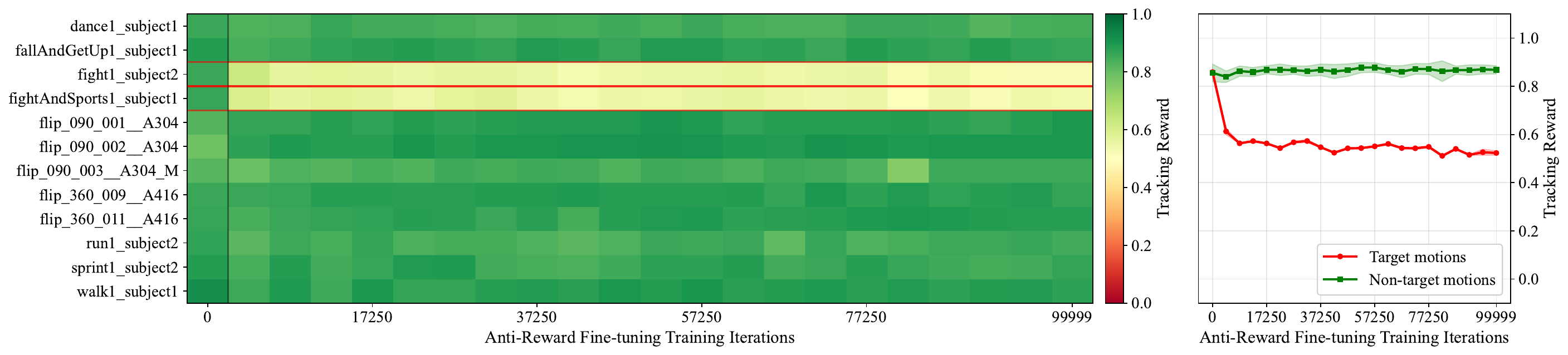}
    \caption{Catastrophic unlearning analysis. The experiment tracks per-clip tracking rewards across 22 checkpoints spanning from the non-unlearning baseline (0) through 99,999 iterations of anti-reward fine-tuning. The left figure shows a heatmap of the rewards for the 12 motions, where the target motions are circled in red lines. The right figure shows the average rewards for the target motions and the non-target motions over training iterations.}
    \label{fig:forgetting}
\end{figure}

Figure~\ref{fig:forgetting} shows that our method does not cause catastrophic forgetting of non-target motions. We track the per-clip tracking reward of all 12 motions throughout the entire unlearning process (99,999 fine-tuning iterations), with a non-unlearning baseline checkpoint (0). The heatmap reveals two distinct patterns: target motions (marked by red boundaries) transition from deep green (tracking reward ~0.86) before unlearning to pale yellow (~0.52–0.61) during unlearning, indicating progressive degradation of the forgotten motions. In contrast, all 10 non-target motions maintain consistently deep green coloring throughout the entire training process, with mean tracking rewards stable in the range 0.84–0.88 and standard deviations below 0.04. Critically, non-target motion performance shows no downward trend—mean tracking rewards at the final iteration (0.868) are comparable to the non-unlearning baseline (0.855)—demonstrating that our method achieves selective forgetting without catastrophic interference on unrelated motions.

\subsubsection{Generalization Across Motion Types Analysis}
\begin{figure}[t]
    \centering
    \includegraphics[width=1\linewidth]{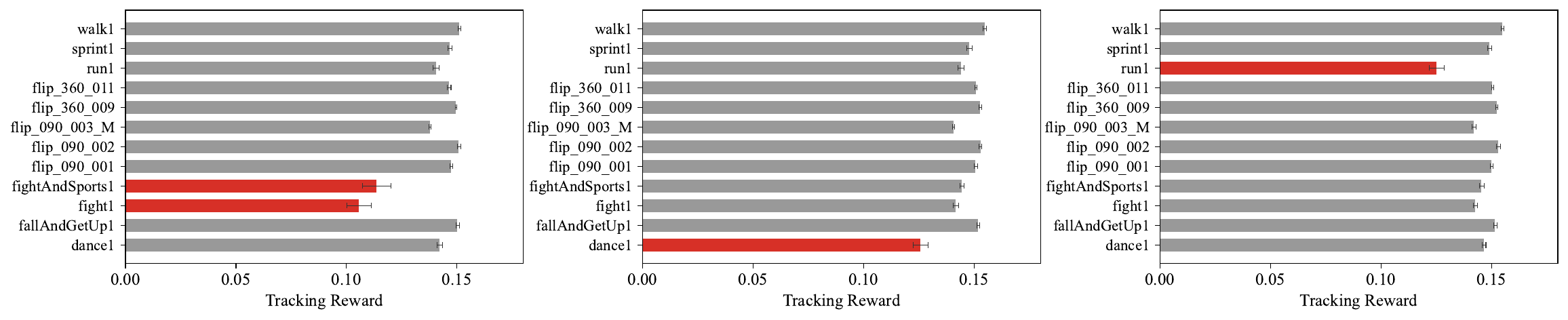}
    \caption{Different motion types as target motions on tracking reward. From left to right, the results correspond to \textit{Fight}\&\textit{FightAndSports1}, \textit{Dance}, and \textit{Run} as the target motions to be forgotten. The red regions are highlighted to indicate that the reward for the corresponding motion decreases.}
    \label{fig:Motion}
\end{figure}

Figure~\ref{fig:Motion} tests whether our method generalizes across motion types of varying complexity. We apply the same unlearning procedure to three distinct motion categories: \textit{Fight}, \textit{Dance}, and \textit{Run}. In each group, one or two motions serve as unlearning targets while the remaining motions measure preservation of non-target motions. The results demonstrate that our method is independent of the target motion to be forgotten and can achieve effective unlearning across different target motions.

\section{Conclusion}
In this work, we present ForgetMimic, the first motion-level unlearning method for RL-driven humanoid control. With the growing maturity of embodied intelligence, removing outdated, inferior, poisoned, and privacy-restricted motions is of critical importance for humanoid control training. As the first unlearning work in this area, our method offers a technical reference for unlearning in the real world and calls for the development of safer, more controllable, and more privacy-compliant humanoid control.

\subsection*{AI use statement}
We used generative AI tools solely for one recommended-disclosure task: creating and modifying scientific figures (Figure 2, Figure 4, Figure 6, Figure 7, and Figure 8). We did not use generative AI tools for any required-disclosure tasks, such as generating synthetic data, developing theoretical models, formulating mathematical claims, assisting in proofs, proposing hypotheses, designing experiments, implementing methods, cleaning or reformatting datasets, supporting qualitative data analysis, or interpreting results. We have reviewed all AI-assisted work, including checking the correctness of the figures. We take responsibility for the final content of this work.


\bibliography{iclr2027_conference}
\bibliographystyle{iclr2027_conference}

\appendix
\section{Appendix}
\subsection{Unitree G1 Visualizations}
\label{ug1v}
\begin{algorithm}[t]
\caption{ForgetMimic}
\label{alg:anti-unlearn}
\begin{algorithmic}[1]
\Require Pre-trained policy $\pi_{\theta_0}$ trained on $\mathcal{M} = \{m_0, \ldots, m_{N-1}\}$, Target motions $\mathcal{C}_{\text{target}}$
\Ensure Unlearned policy $\pi_{\theta'}$
\State Override RSI, Disable assistive wrench, Initialize $\theta \leftarrow \theta_0$
\For{iteration $= 1, 2, \ldots, T_{\max}$}
        \State Sample motion: $m_i \sim \mathrm{Uniform}(\mathcal{M})$, Sample start frame: $t_0 \sim \mathrm{Uniform}(\text{frames of } m_i)$
        \State Execute $\mathbf{a}_t \sim \pi_{\theta}(\cdot \mid \mathbf{o}_t)$, observe $\mathbf{o}_{t+1}, r_{\text{track}, t}$
        \If{$m_i \in \mathcal{C}_{\text{target}}$}
            \State $r_t \leftarrow r_{\text{track}, t} - \lambda \cdot \bar{r}_{\text{track}, t} + r_{\text{reg}, t}$
        \Else
            \State $r_t \leftarrow r_{\text{track}, t} + r_{\text{reg}, t}$
        \EndIf
    \State Compute GAE advantages $\hat{A}_t$ with $(\gamma = 0.99, \lambda_G = 0.95)$
    \State Update $\theta$ via PPO clipped objective (5 epochs, 4 mini-batches)
\EndFor
\State \Return $\pi_{\theta'}$
\end{algorithmic}
\end{algorithm}

\begin{figure}[t]
    \centering
    \includegraphics[width=1\linewidth]{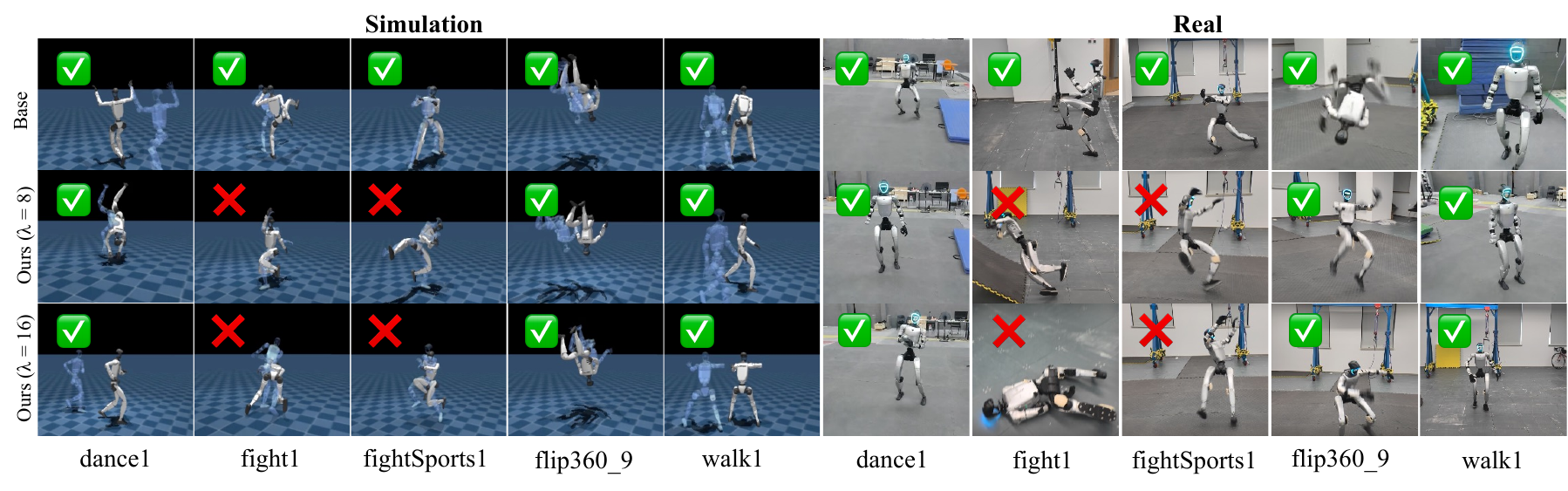}
    \caption{Simulation and real visualizations for the Unitree G1. \checkmark{} indicates the success, and \texttimes{} indicates the failure. The target motions are \textit{Fight} and \textit{FightAndSports1}.}
    \label{fig:visualization}
\end{figure}

Figure~\ref{fig:visualization} shows our simulation experiments on mjlab~\citep{zakka2026mjlab} and real-world experiments on the Unitree G1. We recommend testing in a controlled and safe environment to avoid unnecessary damage to the robot. We summarize our ForgetMimic in Algorithm~\ref{alg:anti-unlearn}.

\subsection{Fine-grained Behavior Quantification Analysis}
\begin{figure}[t]
    \centering
    \includegraphics[width=1\linewidth]{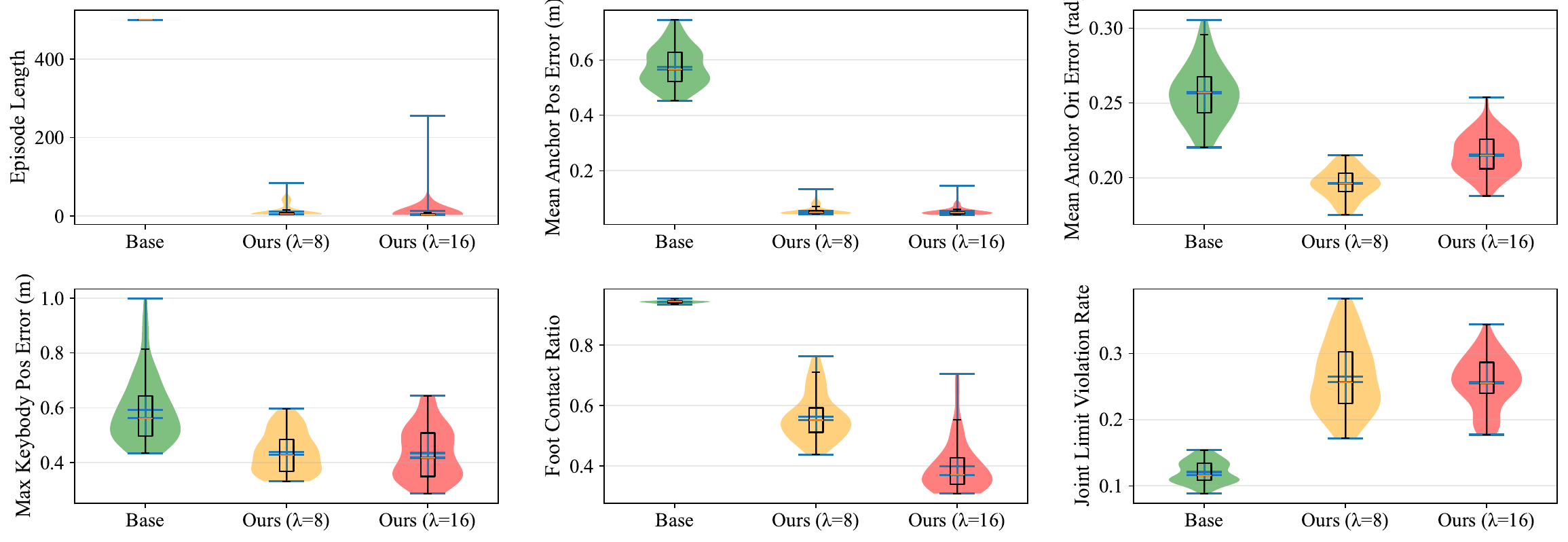}
    \caption{Violin plots for fine-grained assessment of behavioral changes induced by unlearning. Unlike aggregate tracking rewards, these metrics reveal the mechanism of behavioral change.}
    \label{fig:exp10}
\end{figure}

Figure~\ref{fig:exp10} provides a fine-grained analysis of the physical behaviors induced by unlearning on the target motions. We evaluate three setups---Base, Ours ($\lambda=8$), and Ours ($\lambda=16$)---on the two target motions using six physical metrics: episode length, mean anchor position error, mean anchor orientation error, maximum keybody position error, foot contact ratio, and joint limit violation rate. The results reveal a consistent degradation of target motion execution after unlearning. Specifically, the episode length decreases from 500 steps for the Base model to approximately 8--13 steps for the unlearned policy, indicating that the unlearned policies terminate shortly after encountering the target motions. The mean anchor position error decreases from $0.576$ to approximately $0.053$--$0.055$, which is consistent with the policy no longer actively following the reference motion before termination. The foot contact ratio also decreases substantially from $0.94$ to approximately $0.40$--$0.56$, reflecting a pronounced reduction in sustained dynamic motion. In contrast, the joint limit violation rate increases from $0.12$ for the Base model to approximately $0.26$ after unlearning, indicating that the unlearning process can induce physically abnormal configurations when the policy is subjected to the target motions. Overall, these metrics provide complementary evidence that anti-reward fine-tuning substantially alters the physical execution of the target motions rather than merely reducing their tracking reward. The two unlearning strengths exhibit similar trends across most metrics, while $\lambda=16$ produces slightly shorter episodes (7.8 vs. 11.0 steps on \textit{Fight}) and a lower foot contact ratio (0.38 vs. 0.57), suggesting stronger behavioral suppression at the larger anti-reward coefficient.

\begin{figure}[t]
    \centering
    \includegraphics[width=1\linewidth]{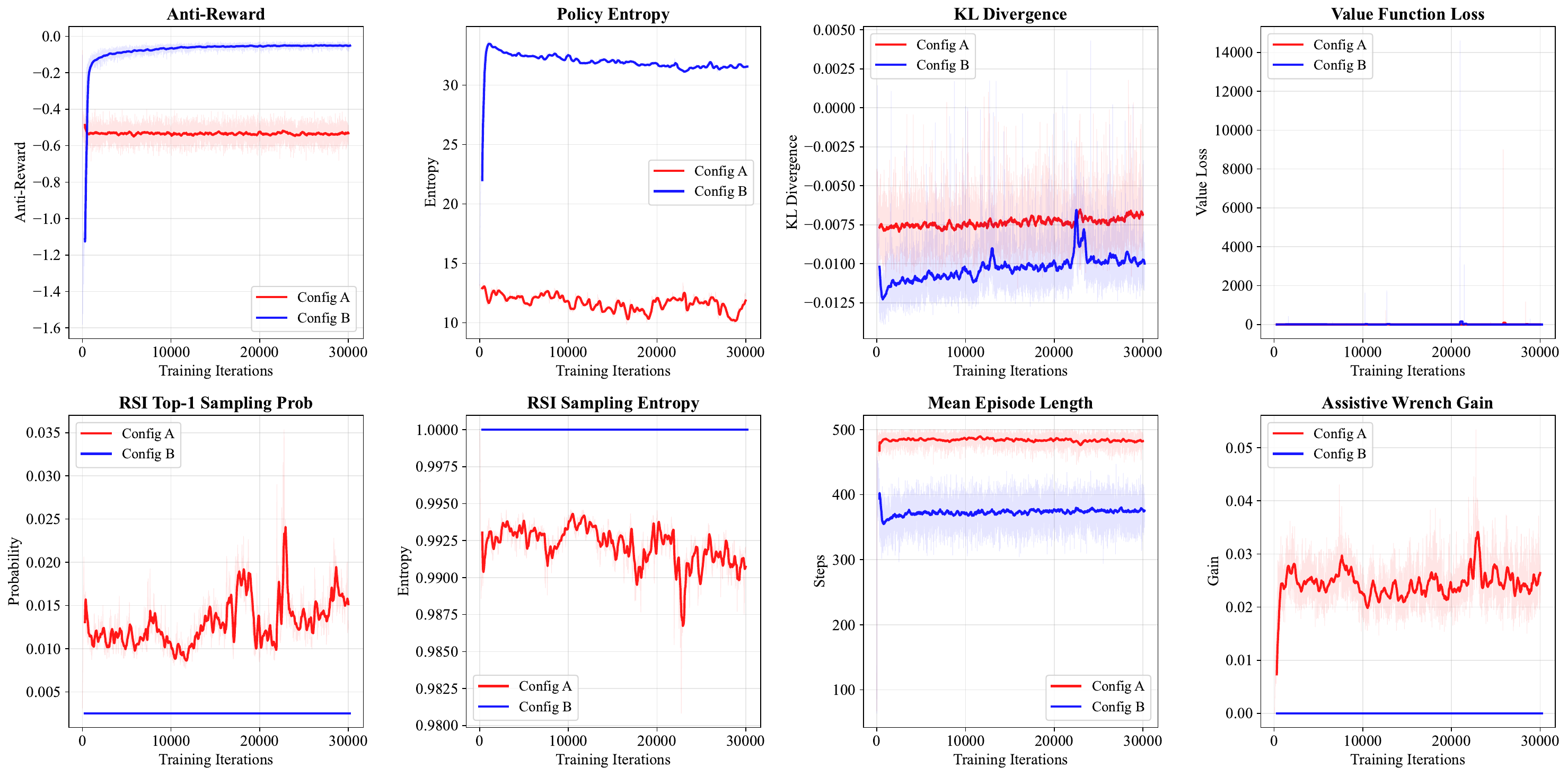}
    \caption{Dynamic behaviors of the two configurations during unlearning training. Config A adopts $\lambda = 2.0$, RSI, and AW, while Config B adopts $\lambda = 8.0$ with both RSI and AW disabled.}
    \label{fig:exp5_training_dynamics}
\end{figure}

\subsection{Training Dynamics Analysis}
Figure~\ref{fig:exp5_training_dynamics} compares the dynamic behavior of two configurations during unlearning training: Config A ($\lambda = 2.0$, RSI, assistive wrench) and Config B ($\lambda = 8.0$, uniform RSI, wrench disabled). From the evolution of eight key metrics, Config B successfully forgets the target motion, whereas Config A exhibits poor unlearning. Anti-Reward ($R_{\text{anti}}$) is the most direct indicator of unlearning effectiveness. For Config B, $R_{\text{anti}}$ rapidly rises from $-1.11$ to $-0.05$ and then stabilizes, indicating that the tracking reward $\bar{r}_{\text{track}}$ on the target motion drops from $0.14$ to $0.006$, close to the random level, so unlearning succeeds. For Config A, $R_{\text{anti}}$ only decreases slightly from $-0.49$ to $-0.53$, and the tracking reward remains around $0.25$, indicating that the policy still retains substantial memory of the target motion. Policy Entropy reveals two fundamentally different learning behaviors. For Config B, entropy increases substantially from $22.0$ to $31.6$, indicating that the policy tends toward random exploration on the target motion, which is a sign of successful forgetting. For Config A, entropy decreases from $12.9$ to $11.9$, making the policy more deterministic, which reflects policy collapse rather than genuine forgetting. Mean Episode Length further supports this conclusion. For Config B, the mean episode length decreases from $394$ to $375$, indicating early termination on some motions, consistent with partial loss of motion ability caused by unlearning. For Config A, the length increases from $468$ to $482$, close to the upper limit of $500$, indicating that the policy can still complete most of the motion sequence. AW gain explains one of the key reasons for the failure of Config A. For Config A, the wrench gain continuously increases from $0.007$ to $0.026$; the physical assistance provides corrective forces when the policy fails on the target motion, directly offsetting the penalty effect of the anti-reward. For Config B, the wrench is completely disabled (gain $= 0$), eliminating this interference. RSI Sampling behavior shows that the adaptive RSI in Config A is active (top-1 probability increases from $0.013$ to $0.015$), but this adaptation instead concentrates training on the target motion and, combined with the incorrect $\lambda$ and wrench assistance, exacerbates the unlearning failure. Config B maintains uniform sampling (top-1 probability remains $0.0025$) and, with the correct penalty strength, achieves stable forgetting. Value Function Loss decreases in both configurations (Config A: $0.047 \to 0.044$, Config B: $0.614 \to 0.194$), indicating that training is overall stable, although Config B has a higher initial value loss, reflecting drastic changes in the reward signal during the early stage of unlearning. In summary, Config B successfully forgets the target motion, whereas Config A fails to unlearn due to multiple issues, including a small $\lambda$, the AW offsetting the penalty, and the adaptive RSI concentrating on difficult motions.

\subsection{Relearning-based Forgetting Evaluation}
\begin{figure}[t]
    \centering
    \includegraphics[width=0.95\linewidth]{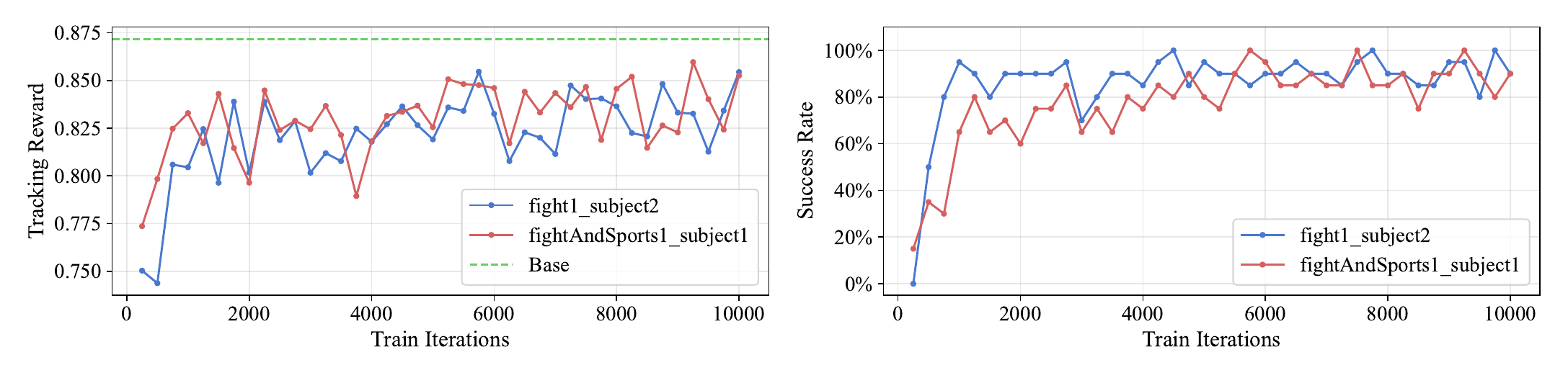}
    \caption{Recovering difficulty on the unlearned policy $\pi_{\theta'}$. The target motions (\textit{Fight} and \textit{FightAndSports1}) are recovered through relearning for the unlearned policy (Ours, $\lambda = 16$). Relearning difficulty measures how difficult it is for an unlearned policy to recover the forgotten motion. A larger number of optimization iterations required to recover the original performance indicates more effective forgetting.}
    \label{fig:exp15_recovery_profile}
\end{figure}

To evaluate the difficulty of re-acquiring forgotten motions, we perform recovery fine-tuning from the $\lambda=16$ unlearned policy for $10{,}000$ iterations using the original positive training rewards. The unlearned policy initially achieves a target tracking reward of only $0.546$, an episode length of $3$ steps, and a success rate of $0\%$, whereas the base policy achieves a tracking reward of $0.8716$ and a success rate of $95\%$. As shown in Figure~\ref{fig:exp15_recovery_profile}, recovering $95\%$ of the base target performance requires approximately $4{,}250$ and $5{,}500$ iterations for \textit{Fight} and \textit{FightAndSports1}, respectively, and approximately $7{,}500$ iterations when recovering both motions jointly. This recovery cost is about $8\times$ the optimization cost of unlearning, indicating that the forgotten motions are substantially harder to re-acquire after unlearning. We therefore use recovery cost as an indirect measure of residual target motion knowledge: a higher recovery cost corresponds to greater relearning difficulty and, consequently, stronger evidence of effective unlearning.

\subsection{Membership Inference Analysis}
To assess whether unlearning reduces membership signals for target motions, we conduct a membership inference (MI) analysis. Given the limited library of 12 motions, we use each target motion as a fixed probe and collect $1{,}200$ rollouts. We extract three features (z-score of normalized tracking reward, episode success rate, and episode length) to infer whether the target motion was learned. Base achieves an accuracy of $0.60$, while Retraining and ForgetMimic achieve lower accuracies of $0.54$ and $0.53$, respectively, suggesting that unlearning reduces target motion's membership. This analysis provides auxiliary evidence of unlearning but does not rule out residual information in local motion fragments or internal policy representations. A full sample‑level MI benchmark tailored for robotic motion data is left for future work.

\subsection{Experimental Details}
\label{ed}
\begin{table}[t]
    \centering
    \caption{Reward terms and their final weights. The reward consists of two major categories: tracking rewards (positive) and regularization rewards (negative or zero).}
    \label{tab:reward_terms}
    \begin{adjustbox}{width=0.9\textwidth, center}
    \begin{tabular}{lccc}
        \toprule
        \textbf{Reward Term} & \textbf{Weight $w$} & \textbf{$\sigma$ / $\sigma_{\text{per}}$} & \textbf{Tracking Target} \\
        \midrule
        \texttt{motion\_global\_root\_pos} & $1.0$ & $0.4$ & Global anchor position \\
        \texttt{motion\_global\_root\_ori} & $1.0$ & $0.5$ & Global anchor orientation \\
        \texttt{motion\_root\_lin\_vel\_b} & $1.0$ & $0.6$ & Anchor linear velocity (body frame) \\
        \texttt{motion\_root\_ang\_vel\_b} & $1.0$ & $1.5$ & Anchor angular velocity (body frame) \\
        \texttt{motion\_body\_pos} & $1.0$ & $0.2$ / keybody (sum) & Relative positions of 14 keybodies \\
        \texttt{motion\_body\_ori} & $1.0$ & $0.4$ / keybody (sum) & Relative orientations of 14 keybodies \\
        \texttt{motion\_body\_lin\_vel} & $0.5$ & $1.0$ & Linear velocities of 14 keybodies \\
        \texttt{motion\_body\_ang\_vel} & $0.5$ & $3.14$ & Angular velocities of 14 keybodies \\
        \texttt{motion\_joint\_pos} & $1.0$ & $0.3$ / joint & 29 joint positions \\
        \texttt{motion\_joint\_vel} & $0.0$ & $2.0$ & Joint velocities (disabled) \\
        \texttt{action\_rate\_l1} & $-0.1$ & --- & L1 penalty on action changes \\
        \texttt{joint\_acc} & $-5 \times 10^{-6}$ & --- & Joint acceleration penalty \\
        \texttt{joint\_limit} & $-1.0$ & --- & Joint position limit violation \\
        \texttt{survival} & $1.0$ & --- & Alive bonus \\
        \texttt{actuator\_torque\_soft\_limit} & $-0.1$ & soft\_ratio $= 0.9$ & Soft actuator torque limit \\
        \texttt{foot\_slip} & $0.0$ & --- & Foot slip (disabled) \\
        \texttt{angular\_momentum} & $0.0$ & --- & Angular momentum (disabled) \\
        \texttt{anti\_shake} & $0.0$ & --- & Anti-shake (disabled) \\
        \bottomrule
    \end{tabular}
    \end{adjustbox}
\end{table}
\subsubsection{PPO Hyperparameters}
We train the policy with PPO using 24 steps per environment per iteration, 5 learning epochs per iteration, and 4 mini-batches. The learning rate is set to $10^{-3}$ with an adaptive schedule, and we use a discount factor $\gamma = 0.99$ and generalized advantage estimation (GAE) parameter $\lambda_G = 0.95$. The PPO clipping parameter is $\epsilon_{\text{PPO}} = 0.2$, the entropy coefficient is $0.005$, and the target KL divergence is $0.01$. We clip the gradient norm to $1.0$ and use a value loss coefficient of $1.0$ with clipped value loss enabled. All models are optimized with Adam.

\subsubsection{Regularization Rewards}
We use the following regularization rewards: an L1 penalty on consecutive action differences (\texttt{action\_rate\_l1}) with weight $-0.1$, a joint acceleration penalty (\texttt{joint\_acc}) with weight $-5 \times 10^{-6}$, a joint position limit violation penalty (\texttt{joint\_limit}) with weight $-1.0$, a per-step alive bonus (\texttt{survival}) with weight $+1.0$, and a soft actuator torque limit penalty with weight $-0.1$ for torque exceeding 90\% of the maximum.

\subsubsection{Rewards Design}
All tracking rewards use an exponential kernel, $r = \exp(-\kappa \cdot e^2 / \sigma^2)$, with $\kappa = 0.25$. The tracking terms and their weights are in Table~\ref{tab:reward_terms}.

\subsubsection{Termination Conditions}
We terminate an episode when any of the following conditions are met: the episode reaches the timeout of 10\,s, the anchor position error in $z$ exceeds 0.35\,m, the anchor orientation error exceeds 0.8\,rad, the end-effector position error in $z$ exceeds 0.25\,m, or the keybody ground contact force exceeds 2000\,N.

\subsubsection{Simulation}
All simulation experiments are performed in MuJoCo~\citep{todorov2012mujoco} using the \texttt{implicitfast} integrator for fast implicit integration. We use a simulation timestep of $0.005$\,s and a decimation factor of $4$, resulting in a control frequency of $1 / (0.005 \times 4) = 50$\,Hz. We set the maximum number of contacts \texttt{nconmax} to $35$ and the maximum number of constraints \texttt{njmax} to $250$. The terrain is a flat plane.
\subsection{Experiments on Unitree H2}
\label{eonuh2}
\begin{figure}[t]
    \centering
    \includegraphics[width=1\linewidth]{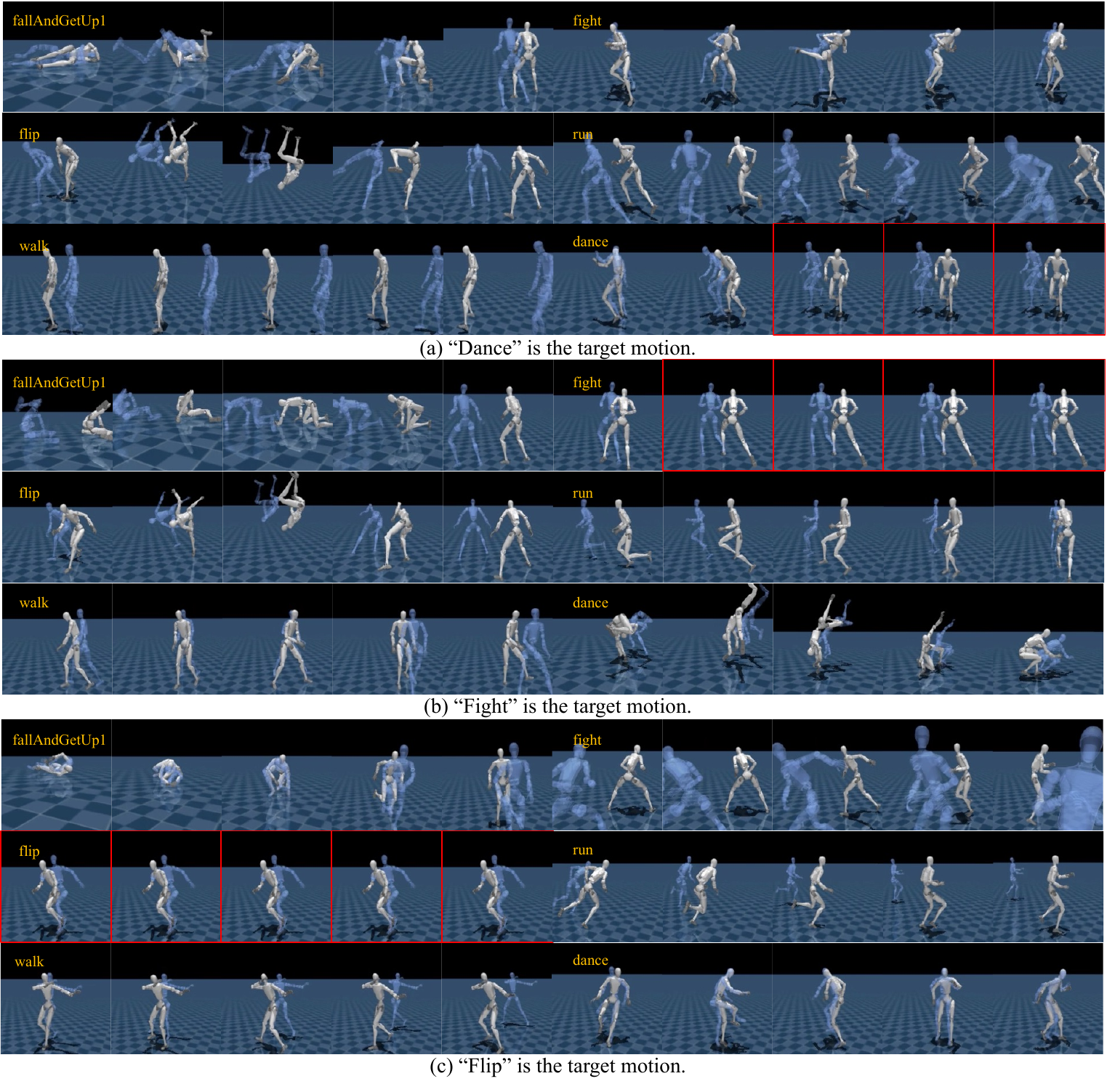}
    \caption{Visualization results of the Unitree H2 in simulation. (a)--(c) show the demonstrations when the target motion is \textit{Dance}, \textit{Fight}, and \textit{Flip}, respectively. Red boxes indicate that the motion episode has already terminated due to joint misalignment. Our algorithm achieves unlearning of the corresponding motion in H2 control while preserving the performance of other motions.}
    \label{fig:h2v}
\end{figure}

\begin{figure}[t]
    \centering
    \includegraphics[width=0.5\linewidth]{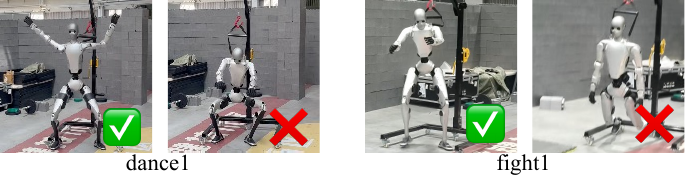}
    \caption{Visualization results of the Unitree H2 in the real world. For the Unitree H2, we conduct the real-world deployment on unlearning for \textit{Dance} and \textit{Fight}. \checkmark{} indicates success, and $\times$ indicates failure. For \textit{Flip}, we do not perform real-world deployment because it can easily damage the hardware. \textit{Dance} and \textit{Fight} are sufficient to validate our method.}
    \label{fig:h2r}
\end{figure}

Table~\ref{tab:h2rs} further evaluates the effect of the number of target motions $K$ on motion-level unlearning. We consider three settings with different target-set sizes: \textit{Dance} ($K=1$), \textit{Fight}\&\textit{FightAndSports1} ($K=2$), and \textit{Flips} ($K=3$). Across all three settings, our method consistently suppresses the tracking ability of the designated target motions. In terms of tracking reward, the target motion performance decreases substantially compared with the Base model, from $0.825$ to $0.681$ for \textit{Dance} ($K=1$), from $0.794$ to $0.654$ for \textit{Fight}\&\textit{FightAndSports1} ($K=2$), and from $0.880$ to $0.631$ for \textit{Flips} ($K=3$). The success rate decreases to $0\%$ for all three target-set sizes, indicating that the proposed method remains effective as the number of target motions increases. Meanwhile, non-target motions, including \textit{Run1}, \textit{Sprint1}, and \textit{Walk1}, retain tracking rewards and success rates largely comparable to those of the Base model. These results suggest that increasing $K$ and different target motions do not substantially compromise the selectivity of unlearning, demonstrating that our method can simultaneously suppress multiple designated motions while largely preserving the remaining motion repertoire. Figure~\ref{fig:h2v} shows the Unitree H2 in simulation, while Figure~\ref{fig:h2r} presents the corresponding real-world unlearning results.

\begin{table}[t]
\centering
\caption{Tracking Reward and Success Rate Comparison on Unitree H2. We consider three motions as the target motion to be forgotten: \textit{Fight}\&\textit{FightAndSports1}, \textit{Dance}, and \textit{Flips}. Underlined values mark the target motion of the corresponding unlearning motion.}
\label{tab:h2rs}
\begin{adjustbox}{width=0.7\textwidth}
\begin{tabular}{lcccccccc}
\toprule
& \multicolumn{4}{c}{Tracking Reward} & \multicolumn{4}{c}{Success Rate} \\
\cmidrule(lr){2-5} \cmidrule(lr){6-9}
Motion 
& Base 
& Fight
& Dance
& Flip
& Base 
& Fight
& Dance 
& Flip \\
\midrule
dance1       
& 0.825 & 0.805 & \underline{0.681} & 0.770 
& 65\% & 60\% & \underline{0\%} & 70\% \\

fallGetUp1   
& 0.846 & 0.842 & 0.831 & 0.820 
& 80\% & 50\% & 85\% & 90\% \\

fight1       
& 0.794 & \underline{0.654} & 0.790 & 0.733 
& 55\% & \underline{0\%} & 55\% & 85\% \\

fightSports1 
& 0.761 & \underline{0.638} & 0.767 & 0.761 
& 85\% & \underline{0\%} & 60\% & 75\% \\

flip90\_1    
& 0.880 & 0.863 & 0.879 & \underline{0.631} 
& 95\% & 95\% & 95\% & \underline{0\%} \\

flip90\_2    
& 0.864 & 0.877 & 0.886 & \underline{0.641} 
& 70\% & 70\% & 75\% & \underline{0\%} \\

flip90\_3    
& 0.843 & 0.841 & 0.851 & \underline{0.715} 
& 90\% & 80\% & 100\% & \underline{0\%} \\

run1         
& 0.777 & 0.792 & 0.801 & 0.759 
& 95\% & 95\% & 100\% & 100\% \\

sprint1      
& 0.832 & 0.817 & 0.809 & 0.783 
& 95\% & 95\% & 100\% & 95\% \\

walk1        
& 0.823 & 0.827 & 0.838 & 0.785 
& 95\% & 100\% & 100\% & 100\% \\
\bottomrule
\end{tabular}
\end{adjustbox}
\end{table}


\subsection{Discussion}

Our method performs approximate unlearning rather than exact, provably certified unlearning, consistent with the setting considered by existing reinforcement unlearning methods~\citep{ye2023reinforcement,gong2024trajdeleter}. In particular, the observed reduction in target motion tracking performance should not be interpreted as a formal guarantee that all information associated with the target motions has been removed from the policy. Instead, our experiments evaluate unlearning from complementary perspectives, including target motion performance degradation, preservation of non-target motions, fine-grained behavioral changes, and the computational difficulty of recovering the target motions after unlearning. The recovery experiments further show that the unlearned policies require substantial additional optimization to restore their original target motion performance, providing indirect evidence that the target behaviors are not readily reacquired.

This distinction is particularly important for humanoid control, where the learned policy contains highly coupled representations shared across multiple motions. Consequently, completely removing information associated with a target motion while preserving all related behaviors may be substantially more challenging than suppressing its execution. Our results demonstrate that ForgetMimic can selectively suppress designated motions while largely preserving the remaining motion repertoire, but stronger guarantees of information removal remain an important direction for future work.

Finally, the proposed formulation is not restricted to a specific humanoid embodiment and can be extended to other reinforcement-learning-based robotic systems. As robots continuously acquire new skills, future unlearning methods will need to account for additional mechanisms that may interfere with forgetting, as well as the potential recovery of forgotten behaviors during subsequent training. Developing theoretically grounded and verifiable unlearning mechanisms for embodied agents therefore remains an important direction for future research.

\end{document}